# A Dual-hierarchy Semantic Graph for Robust Object Recognition

Isaac Weiss

Computer Vision Lab, Center for Automation Research
University of Maryland, College Park, MD 20742
iweiss@umd.edu

**ABSTRACT**

We present a system for object recognition based on a semantic graph representation, which the system can learn from image examples. This graph is based on intrinsic properties of objects such as structure and geometry, so it is more robust than the current machine learning methods that can be fooled by changing a few pixels. Current methods have proved to be powerful but brittle because they ignore the structure and semantics of the objects. We define semantics as a form of abstraction, in terms of the intrinsic properties of the object, not in terms of human perception. Thus, it can be learned automatically. This is facilitated by the graph having two distinct hierarchies: abstraction and parts, which also makes its representation of objects more accurate and versatile. Previous semantic networks had only one amorphous hierarchy and were difficult to build and traverse. Our system performs both the learning and recognition by an algorithm that traverses both hierarchies at the same time, combining the advantages of top-down and bottom-up strategies. This reduces dimensionality and obviates the need for the brute force of "big data" training.

## 1. Introduction

In recent years, the computer vision field has come to be dominated by methods of machine learning (ML) adapted from generic artificial intelligence. These methods do not have much specific understanding of images and rely on extensive training from given examples. These methods are relatively easy to apply and to show some success with, but they do not provide a reliable object recognition solution. The problem is not their immaturity; it is a fundamental limitation due to high dimensionality and nonlinearity, as we shall discuss later. Much more success has been achieved in subdomains such as face recognition and license plate reading where specific knowledge has been applied, but this is hard to generalize. In the following, we describe the problems with current ML methods. We then present our way of applying general shape analysis to provide a general-purpose, inherently robust system for representing and recognizing objects.

The challenges of object recognition stem from the high variability of observed objects. Among the many variables that affect images are viewpoint (pose), scale, illumination, occlusion, articulation, shadows, camouflage, sensor noise, etc. This is in addition to variability in the object itself, before even taking the image, such as when a vehicle is dented or damaged, has added or modified parts, has paint, dirt, stains, etc.

The common methods of machine learning try to combat the variability by training the system with a large set of known images or "templates." Given such training, the system tries to recognize an unknown object based on the training images but without trying to understand the object on a more conceptual level. While this has resulted in success in several areas, the more general problem of variability has not been overcome. There are simply far too many variables to deal with just by the "brute force" method of adding more training.

Another problem with current ML is fragility or brittleness. Changing just a few pixels in an image may be unnoticeable to the human eye but can result in a totally wrong identification by the ML system. Driverless cars have been known to miss a stop sign, misclassifying it as a refrigerator, because it had some (simulated) stickers on it. Of course, one can add training on stop signs with stickers, but it is not possible to train for every possible variation.

It has been recognized [7],[9] that the way to improve the performance of any ML algorithm is to use domain-specific knowledge, namely specific understanding of the particular domain we work on, which in our case is the domain of vision. In other words, we need to advance the study of shape analysis which seems to have fallen out of favor lately. If ML methods outperform more analytic methods in some cases, it does not necessarily mean that we can dispense with shape analysis altogether. It only means that we have to do a better analysis. The success stories in vision involve such specific knowledge in some subdomains. For instance, face recognition under controlled conditions of pose and illumination can be performed now without knowledge about faces, but when trying to go beyond these constraints, it was proved useful to include some modeling knowledge of faces in terms of "landmarks" such as eyes, nose and mouth [14]. The performance of "ignorant" ML methods degrades as we remove more constraints from the query images, and general-purpose recognition is currently out of reach of these methods.

Shape analysis in terms of higher-level visual concepts can improve the recognition performance for general images as well as solve the fragility problem. Moreover, it will give us insight into what the machine is actually doing. With present methods, we have no understanding of how the machine makes decisions or why. This frustrates our intellectual curiosity and defies our long and prodigious experience of applying the scientific method. Even more importantly, we do not want to relegate important decisions to a machine with opaque decision-making. This is especially true in areas such as medicine, defense, finance or law.

Our system applies knowledge pertaining specifically to the vision domain. At the same time, this knowledge is general *within* the image domain, being based on the intrinsic properties of shapes. This enables our method to combine high-level semantics with low-level image processing in both representation and recognition, and to be independent of the source of the image such as EO, IR, LIDAR, LADAR etc., or even CAD drawings.

The contributions of this paper are as follows:

- A dual-hierarchy representation. Many different hierarchies have been tried before to analyze shapes, but they were all one dimensional, such as scale space, DCT etc. We show that such hierarchies are not sufficient to represent even the simplest shapes such as triangles and rectangles. We use two independent but interlocking hierarchies: an object's parts and its level of abstraction. Previous semantic networks typically conflate these two hierarchies into one. These are in a way orthogonal to each other in our method and are traversed differently. Current (analysis-free) ML methods use hundreds of hierarchies that are parallel and are processed in the same way. It is hard to understand what they represent.

- A computational definition of abstraction, independent of human language, and specific to vision. (We use the term abstraction interchangeably with semantics, meaning, generic and sometimes invariance). The definition is based on parts of objects and the geometric relations between them. As these are intrinsic properties of objects, this makes the classification of the objects more robust. Our algorithm can determine the semantics of an object from only a few examples. For instance, in Figure 6 we have circles and line segments in certain positions and orientations relative to each other. When the system sees a few examples of this group, it will give it a name that we can interpret as "face". The smaller circles will be given names that we can call "eyes" and the big circle becomes "head". In this way, the system assigns "meanings" to the circles depending on their context.

- An algorithm of traversing the graph. Previous semantic networks were quite disorganized and difficult to traverse. Our algorithm moves in the two hierarchies simultaneously. Going up in the parts hierarchy (grouping) is similar to a traditional bottom-up strategy, while going down in the abstraction hierarchy is akin to a top-down strategy. We show that this combines the advantages of both while avoiding their pitfalls.

Our two-dimensional hierarchy is rich enough to represent a full variety of objects. Moreover, both our hierarchies represent the intrinsic geometry of objects and their relations with their parts rather than a general parameter such as scale. The parts hierarchy is based on object parts, e.g. a truck has a cabin, a trunk, wheels etc. The abstraction hierarchy includes generic objects on different levels, e.g. a vehicle is more abstract than a truck, a rectangle is more abstract than a door. (This is different from "levels of detail" as we shall see.) Current ML classifications, their version of abstractions, are probably too fragile to build a hierarchy of them. As our definition of abstraction is based on the structure and geometry of the objects, it can be made reliably into a hierarchy.

An abstraction hierarchy is important even in recognizing very specific objects or "fingerprinting". A user wants to know if a specific object seen in the image, say "a blue SUV with a dent on the right door", already exists in the database. The object may be there already, but most likely observed from a different viewpoint and with different illumination. We thus have a high-dimensionality search that current methods have difficulty handling. Our recognition algorithm will go down the abstraction hierarchy to recognize a generic SUV and then recognize the specific dented variant.

Semantic networks have been studied extensively in the context of general artificial intelligence. They derive their semantics from human language or perception, unlike our system. Their generally one-dimensional hierarchy makes no distinction between parts and abstraction hierarchies. Google's "knowledge graph" [6] is probably the biggest. Formal properties of such networks have been studied in [8]. Applications to recognition are described in [1],[2],[3],[4],[5],[13]. They often use graph matching which we do not. An earlier version of the current ideas appeared in [12].

## 2. Limitations of template-based methods

Template matching is the most basic method of trying to recognize objects. Current ML methods share with template matching the property that the training images, or templates, are regarded as nothing more than a collection of pixels, and no attempt is made to analyze the images in terms of higher-level concepts. The templates are classified based on their pixel content. A query image is put into one of the classes based on its pixel content. There are serious problems with these pixel-based methods:

- Huge space of templates. The space of simple binary 100×100 templates has 10,000 dimensions with $2^{10,000}$ possible templates, far more than any computer can ever handle. The "curse of dimensionality" makes it hard to go even beyond 2 dimensions, let alone 10,000.

- Templates representing the same object are not necessarily close in template space. Any classification method based on nearest-neighbor or compactness of classes in a template-based metric (as ML methods are) will fail much of the time.

The large template space manifests itself in practice by large variations of templates, belonging to the same object, resulting from differing viewpoints, scales, illumination, shadows, occlusion, articulation, noise, etc. In addition there are variations in the objects themselves such as a myriad kinds of vehicles, as well as paint, dents, etc. Figure 1 (left) shows some of the possible variations.

Furthermore, some representations of the object are so far from each other that they do not share even one pixel with each other. In the example below, the two representations of the object have the same scale and viewpoint. Yet, they are very far apart in any pixel-wise metric. Humans easily recognize the similarity, but an ML system trained on templates will probably not. This suggests that human visual understanding is not simply a result of training on templates.

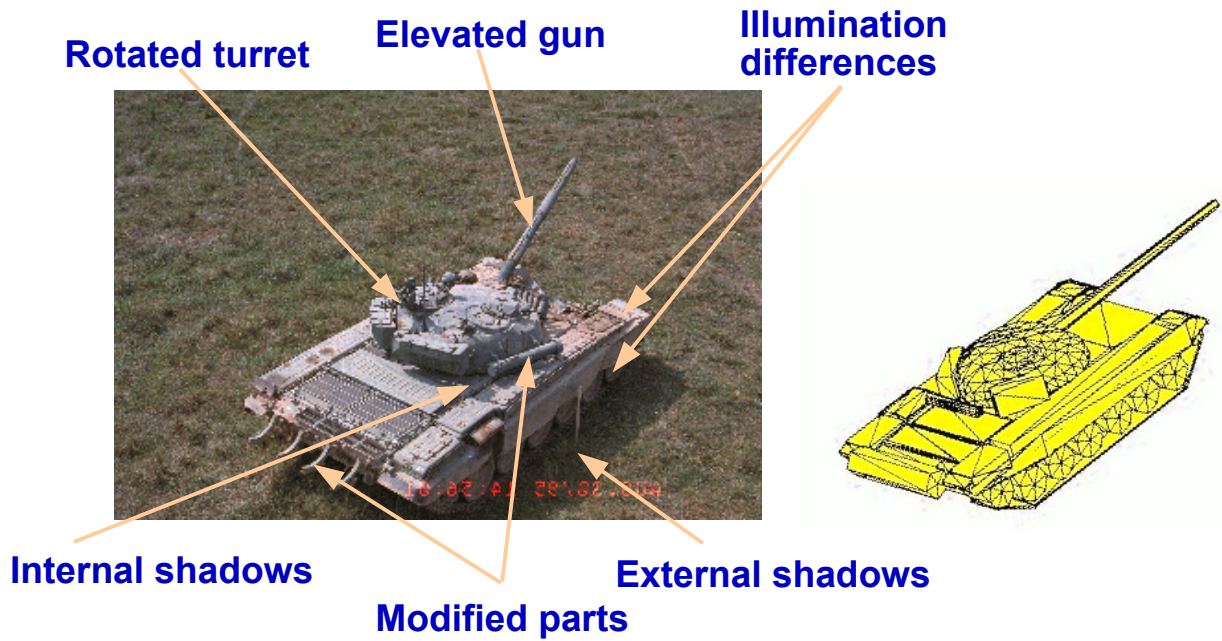

Figure 1: Two representations of the same object.

Machine learning methods try to handle the difficulties in two main ways:

- Brute force: using hundreds of millions of images coming from "big data" databases to train the classifier on.
- Using methods of statistical inference to reduce the dimensionality of the search space and help the classification.

Brute force by itself will never succeed in capturing the full variety of images even with the largest data sets. There are simply too many variables that can change as discussed above. Thus all methods use some kind of statistically-based classifiers.

Most classifiers have some elements in common: a metric that measures distances between class members, a statistical assumption on the probability of these distances, and a nonlinear decision function that affects to which class the object belongs. For example, in line fitting, the metric is the distance of data points from a line, the assumption is that the distances are normally distributed random errors, and a nonlinear function decides which data point is an outlier. Many statistically-based methods have been used in vision: Bayesian networks, support vector machines, principal component analysis, Markov random fields, k-nearest neighbors, minimal description length and many others. The current ML methods also contain the above elements in one way or another.

These methods are quite generic, and they do not have much understanding of the structure of objects or images. They typically use a pixel-wise metric and assume some random error in this metric. However, looking at Figure 1, it is evident that the pixel-wise distance between these two pictures is not just random error. There are systematic differences there that cannot be captured statistically. The various variables involved influence the image in a nonlinear way. Shadows or gray levels depend nonlinearly on the positions of several unknown light sources, projection from 3D depends nonlinearly on an unknown viewpoint. Thus, the distance between objects is very different from the distance between the corresponding images. Objects can be close in the object space and far apart (pixel-wise) in the image space, and vice versa. The boundaries of the classes differ greatly in the two spaces in a complicated nonlinear way. Current ML methods try to bridge this difference by a combination of linear smoothing filters and generic rectifying functions. This often fails in unexpected ways such as overfitting, or amplifying differences between objects when there are almost none in the images.

We can improve on this basic method by applying some knowledge about images. In the example of Figure 1, we know that edges matter more than pixels in recognizing objects, so we can use an edge detector on the real image on the left. This would make it much closer to the drawing on the right. Of course edge detectors have their own problems so we need to apply further knowledge like corner detection, points of interest, etc. to make the images closer.

This example is a special case of a general theorem that characterizes the performance of ML methods as applied to various problems. It is known as the NFL (No Free Lunch) theorem [7],[9]:

**Theorem (NFL): Any two search algorithms are equivalent when their performances are averaged over all possible problems.**

This means that we are unlikely to improve performance by tweaking some ML algorithm that was applied before or using a variant of it. Over a large set of problems, the performance levels of all these algorithms will converge to the same level. Given the difficulties of ML in high-dimensionality nonlinear problems, this performance level leaves much room for improvement.

In other words, we have to "earn" our lunch and improve performance by using specific knowledge about vision rather than generic statistical assumptions. We cannot take the easy route and rely on a statistical machine learning method to do the task of image understanding. We have to use methods of shape analysis to obtain a true understanding of the structure and relationships of images and objects. Although ML methods outperformed systems that use analytic knowledge in some cases, this does not mean that we can eschew knowledge of vision altogether as is the current trend. It only means that we have not developed and applied this knowledge correctly.

In the following we will apply knowledge of the shape and structure of objects and images to recognition and learning. Our distance metric is based explicitly on the features and structure of the objects. Our nonlinear decision function is based on positive feedback between the object, its parts, and other related objects.

## 3. The dual hierarchy

A good way to reduce the dimensionality of a search is to use a hierarchy. Many kinds of hierarchies have been tried: scale space, wavelets, "pyramids", quadtrees, Fourier transform, discrete cosine transform, parts decomposition, levels of detail, semantic levels, etc. The idea is that the higher levels of the hierarchy have lower dimensionality so recognizing objects becomes much easier. Once an object is recognized at a higher level, the information can be used to perform a more refined recognition.

All these methods have one problem in common: they are one-dimensional hierarchies. That is, there is only one parameter that changes as we go up the hierarchy, such as scale or level of detail. However, one dimension is not enough to represent the wide range of possible objects with all their variability. A simple example can demonstrate this. A polygon can be decomposed into parts, namely its sides. In a parts hierarchy the sides are lower than the polygon. The number of parts can distinguish different polygons from each other, e.g. a triangle, a quadrilateral, or a pentagon. However, it cannot distinguish a generic quadrilateral from a rectangle, and a rectangle from a square. These are distinguished from each other by an independent hierarchy, involving geometric relations between the parts of the object. A rectangle is more specific (less generic or less "abstract") than a quadrilateral, as its angles are specified as all equal, and a square is more specific than a rectangle, as all its sides are also equal. We can see that both of these hierarchies, the parts and the abstraction, are essential for describing even these simplest of objects. The hundreds of hierarchies used in current ML are obviously redundant for this example.

More real-life examples abound. A generic, or "abstract" vehicle is composed, in the parts hierarchy, of generic parts such as wheels, a body, doors, windows. A generic wheel is composed of a rim, a tire, a hub, spokes, etc. In the abstraction hierarchy, a generic vehicle has more specific vehicles on a lower level of abstraction, such as a car, a truck, or an SUV. These have similar parts but with different geometric relations. Similarly, a car can be further specified as a sedan or a sports car etc. Generic wheels can be specified into types on a lower level of abstraction by geometric properties such as the thickness of the tires or the shape of the spokes. Thus, the two-dimensional hierarchy of parts and abstraction is quite sufficient to represent quite complex examples like these.

In the following, we generalize these examples and use a dual hierarchy for our representation of objects:

- A parts hierarchy, in which objects are composed of simpler parts and those are made of yet simpler parts.
- An abstraction hierarchy, in which "abstract" also means "generic" as opposed to specific. This is independent of the parts hierarchy. A rectangle has the same parts as a square but is more generic.

Both the abstraction (as defined below) and the parts hierarchies derive from the intrinsic nature of the object itself rather than from some general parameter such as scale or frequency. Thus, these hierarchies are invariant to various changes in the environment, such as changes in illumination or viewpoint, as well as changes in the objects themselves such as paint or dents. Non-intrinsic parameters such as scale are not only not invariant but can distort the shape. For instance, a scale space hierarchy is generated by smoothing an image by (e.g.) a Gaussian filter. A limited amount of smoothing may be useful for some noise reduction, but when taken too far it totally distorts the shape. A rectangle will lose its corners and gradually morph into a circle, an unrelated shape, as we go up in scale space.

## 4. What is an abstract object?

Abstraction, as philosophers have pointed out, is the process of finding generalities among specific things, or finding quantities that are invariant among several specific objects. (The most specific objects in our case, at the lowest level of the abstraction hierarchy, are 2D or 3D images.) Such abstract objects can be further abstracted at a higher level. Thus, an abstract object can be defined as a class of objects, specific or abstract, with common properties. The abstraction process is usually done by humans using human perception and language. Here we use abstraction principles that can be applied computationally without human intervention. This abstraction is based on the intrinsic structure and geometry of the objects, so it makes the classification more robust. That is, the similarity or distance between objects is defined based on their structure and geometry, not on pixels. We obtain such abstractions by several means:

- **Geometric relations.** Looser geometric constraints mean higher abstraction, as they are common to a larger set of objects. Given a set of specific triangles, we can drop the specific lengths or angles of the sides and only keep the property that it has three touching sides. This property is common to all specific triangles and is invariant to the lengths, thus defining an abstract triangle. This is independent of the parts hierarchy as all triangles consist of three sides as parts. In the example of the rectangle and square above, we used the geometric relations between parts, such as their relative sizes or angles. The square requires equal sides and angles. Dropping this specific requirement, we obtain a rectangle, which only requires equal angles and is thus more abstract. In Figure 6 (right), an arrangement of circles and line segments is seen as an abstract "face". This happens because reasonably loose geometric relations between those objects are common to most faces.

- **Context.** The abstraction or "meaning" can change according to the group the object is found in, namely the context. A circle can be an eye in the context of a face or a wheel in the context of a vehicle. In Figure 4, using knowledge specific to the face group, the big circle acquires the specific meaning of a "head", the small circles are now "eyes" and the line segments are a nose and a mouth. Thus these parts acquire specific "meanings" through the context of a group they are parts of. Both the eye and the wheel are abstract, generic objects but at a lower level of abstraction than a circle, as they are more specific than it. We treat a circle and an eye as separate objects in the hierarchy even when they are related to the same set of pixels in the image, because the eye is a part of the face group while the circle is not.

  In the simple rectangle example, Figure 5, we first recognize the rectangle by the relations between line segments in the image. Given that, these line segments become "sides" of the rectangle, namely more specific objects than line segments, through their being parts of the rectangle. The rectangle can become a "window" in the context of a house, namely a group containing several rectangles, or a face of a box as a part of a 3D box.

- **Generic parts.** An object having generic or abstract parts implies an abstract object. In Figure 4, the generic face is made of generic eyes, mouth, etc. A truck can have a generic cab and a generic trunk as parts. Changing the kind of cab will not change its nature as a generic truck. This is illustrated in Figure 3. The generic truck (top left) is made of the generic cab and trunk (top right). The generic cab is connected to more specific cabs (bottom), which the generic truck does not know about.

- **Levels of Detail.** This is only one source of abstraction, commonly conflated with both the abstraction and the parts hierarchies. Some parts of the object are less important in recognizing it, so they can be omitted in a more generic version of the object. A chair has the essential parts of legs and a seat, while an armchair also has arm rests. Thus, a chair is higher in the abstraction hierarchy than an armchair. This is independent of the parts hierarchy. A chair is not an abstraction of its legs and seat; it is a group higher than they are in the parts hierarchy.

- **Variable parts** such as the unspecified number of sides in a generic polygon.

In all the cases above, the parts and abstraction hierarchies are independent but complementary to each other, and they interlock and intersect with each other. We could not have distinguished different objects such as a circle and a (schematic) eye without both the abstraction and the parts hierarchy.

## 5. The graph representation

The question immediately arises of how to represent the dual hierarchy of parts and abstraction. Parts decomposition is quite easy to represent. However, how do we represent an abstract object such as a generic triangle? This seemingly simple problem has occupied philosophers since Plato. No one has ever seen a generic triangle—all the triangles we have ever seen are specific, having specific values for the lengths of the sides. Yet, we do have a concept in our minds of an abstract, generic triangle. Obviously we have the verbal definition of a triangle, but this will not work for more complex objects such as vehicles. All the vehicles we have ever seen are specific vehicles, but we do have a concept of a generic vehicle. An exact verbal definition of a generic vehicle seems impossible for such a complex and variable object. Philosophers have argued for millennia whether abstract objects even exist in the world or they are only concepts in the mind (the "problem of universals"). In our method, an abstract object exists as a node in a graph, connected to nodes of more specific objects. The generic triangle is depicted graphically in Figure 2. A truck example is in Figure 3.

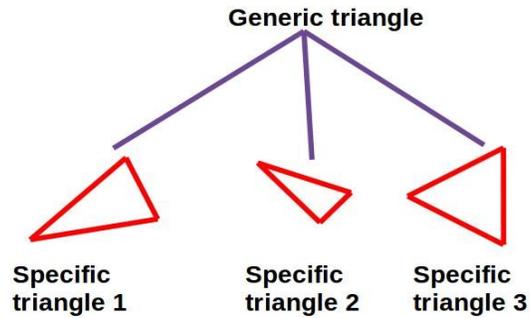

Figure 2: Representing a generic triangle.

In the following we describe a graph, or a network, that represents generic as well as specific objects with all their variations. This graph is organized as a dual hierarchy that represents both our parts and abstraction hierarchies. All objects, specific and generic, are represented as nodes in the graph. The relations between nodes are represented as links in the graph. Various graph and network methods have been tried before, but they never had the clear distinction between parts and abstraction hierarchies that we have here. We do not presume to know if biological vision systems are organized this way, but it is easy to imagine that such a network can be built from neurons.

A simple example is shown in Figure 3. In this figure, a generic truck and two specific trucks with their parts and all known variants are represented in the same graph. The parts hierarchy is shown here left-to-right, i.e. a truck is made up of a cabin and a trunk as seen on each horizontal level. The abstraction hierarchy is shown vertically. At the top-level of this hierarchy, a generic truck node is linked horizontally (i.e. in the parts hierarchy) to a generic cabin node and a generic trunk node. Similarly, on the lower abstraction level, the specific "truck1" and "truck2" are linked horizontally to the specific nodes of their respective parts, namely specific cabins and trunks. Our generic truck node is linked vertically to both the specific "truck1" and the "truck2" nodes lying at the lower abstraction level. Similarly, the generic parts, the cabin and the trunk at the top abstraction level, are linked vertically to more specific nodes of cabins and trunks at the lower level. (Objects that are at the same position in both hierarchies are arranged on diagonal lines that can represent a depth dimension. They are not linked to each other.)

A specific instance of an object can also be seen as a small subgraph that includes the node of the object, e.g. "truck1" (Figure 3), its specific parts' nodes, and a node for a corresponding generic object. This helps represent variants of an object. While each known object is represented by a node, variants of the same object do not need a separate node. We can have an object "truck1" (Figure 3) with two variants, containing either "cab1a" or "cab1b" as a part. The subgraph highlighted in the figure contains the parts of the specific variant. Thus, the graph can represent many more variants of objects than the number of nodes it contains. (If we choose to we can separate the variants into different objects with specific nodes.) This is useful as it would be quite hard to account separately for every possible variant of every object.

This example shows that the graph provides a very flexible definition of objects. We are not restricted to one definition of the "truck1" but we accommodate different variants having different variant parts, and we can also place them all as subgraphs under the generic truck. Thus, a large class of specific objects can be organized very economically as subgraphs in the dual hierarchy under the same generic object. At the same time the graph remains understandable.

A general model graph includes objects from the very simple such as edges, corners, or line segments, to mid-level features such as rectangles or circles, to the highest levels of objects to be recognized, each object having a position in each of the two hierarchies. In this way, we integrate high-level knowledge about the structure and geometry of objects with low-level data.

Another example, Figure 4, illustrates context-dependent abstraction. A circle can be an abstraction of a wheel, an eye, or a head, all having different semantic meanings derived from the group they belong to. Thus, the node "circle" is connected to the "eye_i", the "head" and the "wheel" nodes which are at a lower level of abstraction. The eyes and head acquire their meanings by being parts of a face, while the wheel becomes so by being a part of a truck. Accordingly, the "eye_i" and "head" nodes are connected to the "face" node higher in the parts hierarchy, while the "wheel" node is connected to the "truck" node. The circle node is not connected directly to the face node, as different circles in the face have different meanings (eyes, head). All these "circular" objects are assigned different nodes in the graph, and are distinct from the corresponding sets of pixels in the image. (Obviously they are all more abstract than these pixels.) Thus, our graph affords us a very flexible representation of semantic knowledge as defined in Section 4.

The "circle" node represents an abstract circle and not the perfect circles shown in the drawing. It is impossible to draw a generic circle, as all drawn circles are specific. The generic circle can be connected to more specific circles with various distortions, gaps, or noisy data.

In addition to the objects, we need to represent the relations between objects and/or parts. These are represented by the links between the nodes. The relations can be geometric, e.g. distances, angles, relative sizes of parts etc, or they can be topological, e.g. a part touches another one or is contained within it. Topological relations are inherently invariant to geometric transformations such as viewpoint changes. We want our quantitative geometric relations to be invariant too. Distances and angles are not generally invariant, so we replace them as much as possible with invariant properties. These include parallelism, symmetries, ratios of lengths, and more general algebraic invariants. A graph is said to be "attributed" if the links between the nodes possess some quantitative properties, or "attributes". Thus, our representation consists of an attributed graph containing object nodes and their attributed links.

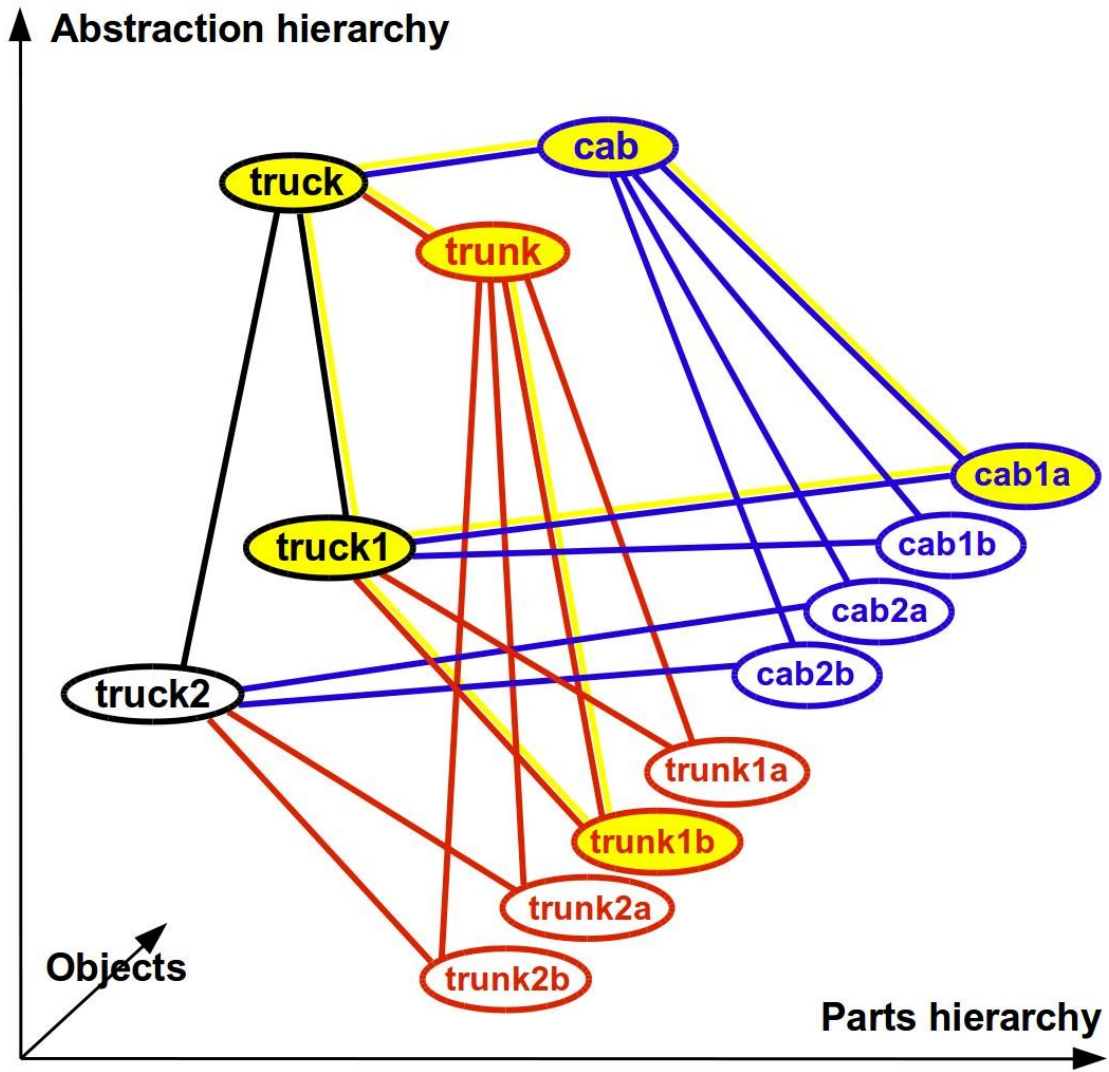

Figure 3: The dual-hierarchy graph. Recognized objects are highlighted.

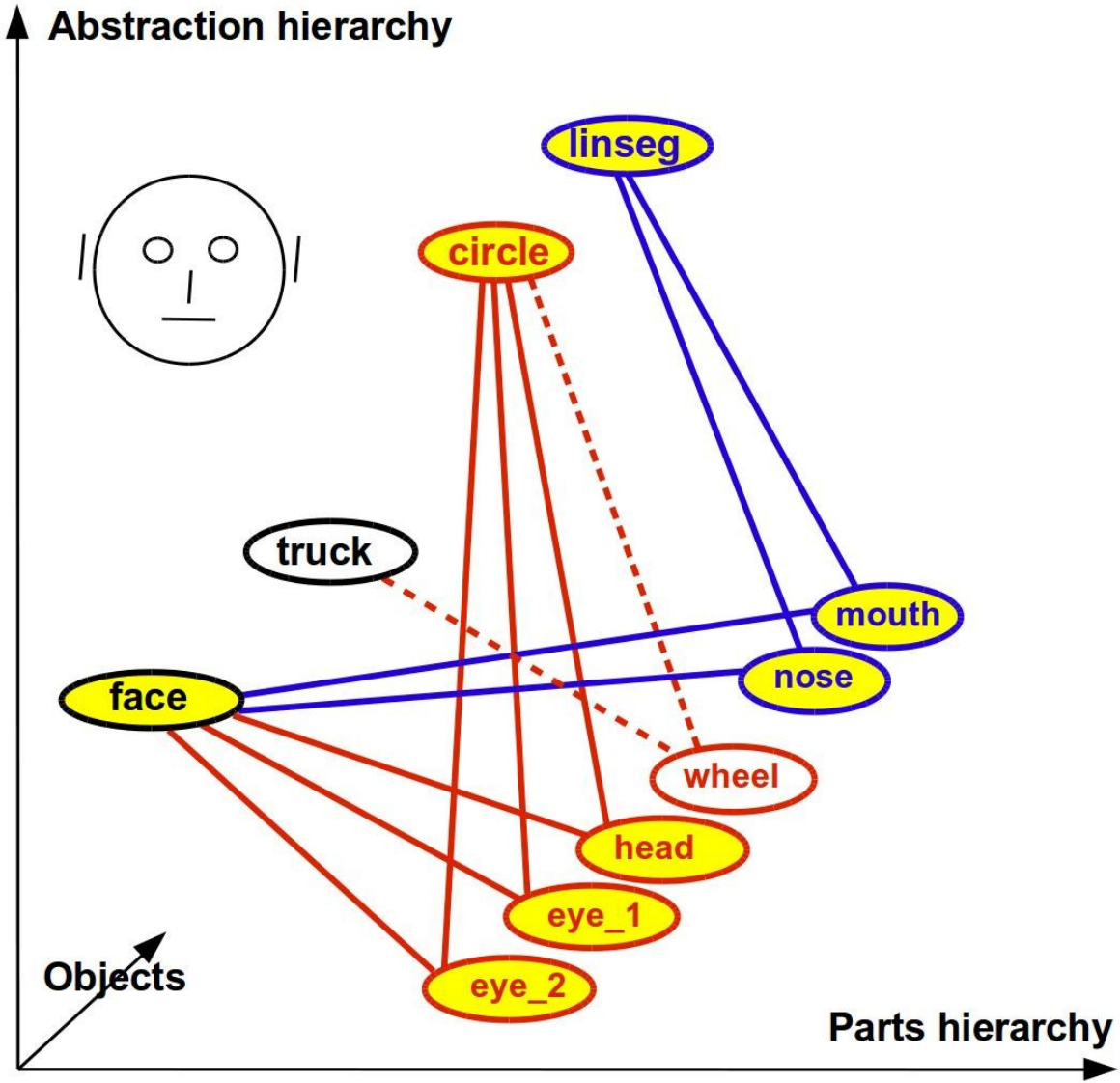

Figure 4: Model graph of a face.

The graph representation, being intrinsic to the objects, is invariant to external influences such as viewpoint or illumination changes, shadows etc. It is also invariant to the platform – whether we use EO, IR, LIDAR, LADAR or a CAD model so we can use the same representation. Thus, it can be used as a form of a unified dataset of objects across all modalities. This can be useful for unifying the existing datasets of so-called "semantic levels". These are related to, but not the same as, our levels of abstraction. These are currently separate datasets, e.g. a dataset for raw data, a dataset for polygon models, etc. We can represent them in a unified way in the same graph.

## 6. The recognition algorithm

The graph containing all models is constructed offline and serves as our dataset of models. Given an object to recognize, we now need to search for the correct model in the graph hierarchies. The dimensionality of the search is reduced by the two hierarchies.

Traditionally there are two competing approaches to traversing hierarchies: top-down and bottom-up, each with its own serious problems. Starting at the bottom, there are many raw features such as edges, and we face a combinatorial problem of how to group them in the right way to obtain the higher-level object. This is greatly compounded by noise and uncertainty in the features. Trying to start from the top, we need to somehow guess which high-level object we might have and try to fit it to the low-level features. We do not often have such a guess. In addition, it is not always clear if an object is a high- or low-level object. Given a circle, it can represent a whole object such as a head, or a small part such as the pupil of an eye.

Our dual hierarchy makes it possible to use both strategies at the same time, combining their advantages and avoiding their problems. In a nutshell, we use a top-down traversal in the abstraction hierarchy and bottom-up in the parts hierarchy. A circle is both a low-level primitive in the parts hierarchy, e.g. the pupil, and a high-level abstraction in the abstraction hierarchy, such as a generic head. Thus, a circle can be a starting point for both hierarchies (Figure 7).

Another problem with traditional methods is that they try to build an image graph separately from the model graph and then try to match the two. For example, they try to find edges, lines, circles, etc. with no knowledge of what the object is and then try to match these parts to the model. This often fails because the visible lines or circles are often ambiguous and cannot be reliably detected until the object itself is recognized. In our algorithm, building the image graph is directed at each step by knowledge from the model graph and not by some generic heuristics. Object recognition thus amounts to building the image graph with guidance from the model graph.

The recognition algorithms proceeds using two processes:

- **Generation of grouping hypotheses:** bottom-up in the parts hierarchy. Given some detected objects in the image, with detected geometric relations, we use them as "clues" to find a suitable hypothesis. We typically only need two such objects. We look them up in the model graph to see if there is a group there that contains these objects as parts with the given relations. For example (Figure 5), if we detect two line segments that touch each other at a right angle, we check the model graph and find the group "rectangle", which contains such line segments as parts. This is a group hypothesis. In the example of Figure 6, the clues are a pair of circles and a line segment in certain relations, which lead to the hypothesis of a face, with a suitable coordinate transformation. Of course, this is only one hypothesis and we need to check others. The number of checks can be reduced, e.g., by first checking for common intermediate combinations such as "circle_pair".

- **Verification and specification:** top-down in the abstraction hierarchy. For a hypothesized group, we look up its parts in the model graph and check if there are corresponding objects in the image. If so then the hypothesis is verified. These parts typically have meanings within the group which are on a lower level of abstraction than the corresponding objects we already found in the image, so we add them to the image graph under the corresponding objects we found. In the above hypothesis of a rectangle, we find in the model graph that a rectangle has four parts, called "side"s, with certain relations and so we look for them in the image. Finding four corresponding line segments verifies the rectangle hypothesis. We add the node "rectangle" to the image graph at a higher level than the "linseg" nodes (line segments) in the parts hierarchy, with appropriate connections. The "side" parts of the rectangle are more specific objects than line segments and so we add a node "side" to the image graph under each corresponding "linseg" node on the lower level of abstraction. In the face example (Figure 7), we find that the "face" parts corresponding to circles in the image are "eye_i" and "head", so we add these nodes under the corresponding "circle_i" nodes in the image graph. The new nodes are assigned probabilities in an algorithm described in the next section.

We iterate similarly to the next levels. We can see from these examples that the image graph is built incrementally using knowledge stored in the model graph. As the recognized objects are represented by the image graph, building the image graph constitutes recognition. No graph matching is involved.

Figure 5 illustrates the iterative process for a truck. The red lines represent the hypothesis generation and the green lines represent the verification. Dotted lines represent failed hypotheses such as a "square". Successful hypothesis nodes are highlighted in yellow and are copied to the image graph. As described above, we start from the "linseg" nodes (line segments) and hypothesize the "rectangle" node with the grouping process. The verification process verifies this node and creates the "side" nodes under the "linseg" nodes, with appropriate connections, as the sides are more specific than line segments. The rectangle is obviously higher in the parts hierarchy than its sides and it is at the same level as they are in the abstraction hierarchy. In this way the algorithm moves one step up in the parts hierarchy and down in the abstraction hierarchy.

In the next iteration, we apply the grouping process to the rectangles and create the "box" hypothesis. (There can be many rectangle nodes in the image graph while there is only one in the model graph.) The 3D projection is taken into account when checking the relations between the rectangles that make up the boxes. The verification process finds the "box-face" parts of the box and places them under the corresponding "rectangle" nodes.

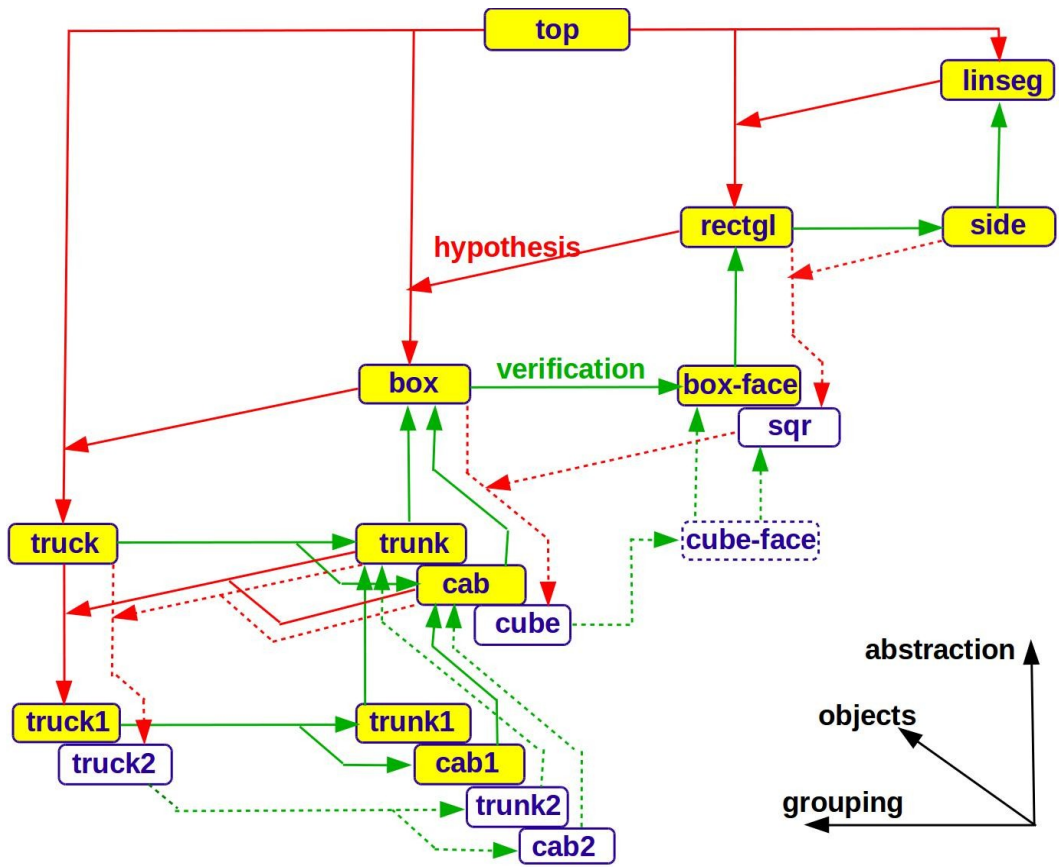

Figure 5: The traversal algorithm.

We proceed similarly to more interesting objects. In the next iteration, we check if the boxes have the right geometric relations to group them into a "truck". The verification and specification process then find the "cab" and the "trunk" parts as specific meanings of the corresponding boxes that generated the truck hypothesis, so these nodes are added under these box nodes. The node "truck" in the model graph is linked to additional parts such as wheels and bumpers, so we try to locate them in the image. A "wheel" node is then placed under a "circle" node in the abstraction hierarchy if a circle is present in the image in the right place. This further verifies the "truck" hypothesis. In the next iteration, we use more specific relations between the cab and the trunk of the truck to find the more specific "truck1". We add this node to the image graph. The failed hypothesis "truck2" is not added.

The relations used for hypothesizing are stored in the relevant group. For example, the specific relations for hypothesizing truck1 and truck2 are stored in the generic truck and do not need to be checked when looking for specific faces. The initial hypotheses relations are in the "top" node.

The face example is shown in Figure 7. We start from two similar circles and one line segment found in the image. Given their geometric relations, we hypothesize a face in the model graph, Figure 4. Figure 7 shows the image graph that we built. The eyes, mouth, nose and head were verified in the image, so the face was verified. The ears were not verified.

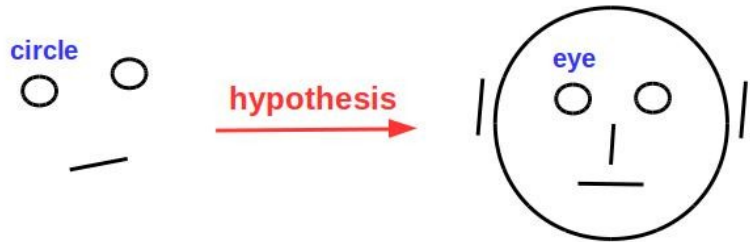
Figure 6: Face hypothesis.

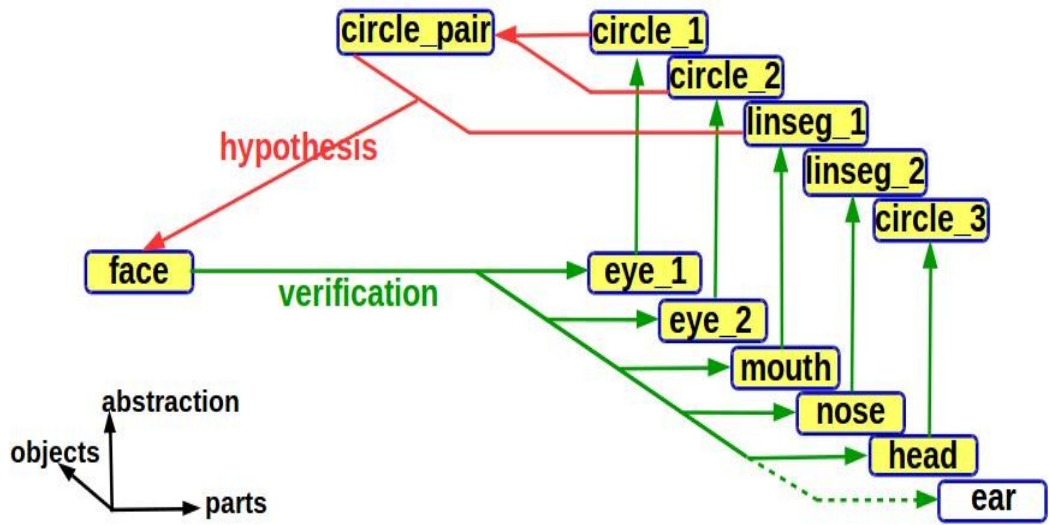
Figure 7: Face recognition – building the image graph.

In summary, we go up in the parts hierarchy and down in the abstraction hierarchy at the same time, simultaneously grouping parts and recognizing more specific objects. This process is directed through all levels of the processing by prior knowledge, stored in the model graph hierarchies. This avoids errors such as noise and shadows because these are not present in the model graph.

## 7. Propagation of probabilities

Obviously, the relations expressed in the model graph will not be satisfied precisely in the image. Thus we need some measure of a distance between the ideal model and the image. This determines the probability of the match. Unlike pixel-based methods, we measure this distance in terms of the nodes of the graphs rather than pixels. This takes into account the structure and geometry of the objects.

Each node we create in the image graph is assigned a probability for its match to the model graph. The probabilities propagate through the graph in a way somewhat similar to a Bayesian network. However, unlike a Bayesian network, we allow some propagation backward. This can create feedback loops, so we limit the backward propagation in a way that prevents this. Thus the probability of an object depends on its structure, i.e. not only the object's node itself but also the probability of the other nodes connected to it, as in the subgraph in Figure 3.

Each node in our graphs has a coordinate system with an origin and principal axes. The axes represent the orientation and dimensions of the object. These are the variables in the geometric relations functions between the nodes, such as ratios of distances. All these relations are assigned elasticity or "spring" constants that represent the tolerances of these relations. We use the elastic strain as a measure of the conditional probability of one node to match given the others. Obviously, the conditional probability is lower when the deviation of the relations from the model is higher. Maximizing the probability means minimizing this deviation, or our distance measure, of the image graph from the ideal model.

Each step in the recognition algorithm above creates a bunch of new nodes and links. We update the probabilities of nodes in an iterative process that takes into account the probabilities of all nodes connected to a particular node, as well as the conditional probabilities in the links. Only the most elementary nodes such as edges are connected by "springs" to the raw image data. All nodes can move on their connected springs. The goal is to update the positions and axes of the nodes so that the overall probability is maximized. We proceed as follows.

- For every new node created in the image graph, sum the probabilities contributed from all the nodes linked to it, namely the other nodes' probabilities multiplied by the conditional probabilities expressed in their links.

- For every old node connected to a new one, update its probability using a similar sum, this time including the new nodes. Then, proceed to update other nodes connected to that node (with limits). Drop weak nodes.

- Links (conditional probabilities) are updated, and weak links are eliminated. This allows, say, in a scene of a crowd, to decide if a hand belongs to the person on the right or on the left of it, keeping only the strongest links.

- For each new or updated node, adjust its position and axes to maximize the overall probability.

- Iterate until convergence.

This algorithm provides a nonlinear decision function that is based on the intrinsic structure of the object. The probability of a truck in the image is increased when the cab and the trunk are strengthened, e.g. by finding wheels, and vice versa. A line segment that looks faint in the image may be interpreted as a side of a rectangle. The connections of this rectangle to other sides strengthens this interpretation, and this in turn further strengthens the rectangle. When the faint line segment is not connected to anything, its interpretation is weakened and it is deemed noise. This mutual influence between nodes is useful, as long as it is supported by other connected nodes, and feedback loops are avoided.

## 8. The learning algorithm

Since our definition of "abstraction" does not depend on human language, our learning algorithm can be fully autonomous. The algorithm uses its own rules of abstraction to learn from sample images, as discussed in Section 4. Our learning algorithm is analogous to the recognition algorithm described in Section 6: we go up in the parts hierarchy by grouping simpler parts, and simultaneously, we go down in the abstraction hierarchy from generic to specific objects. Using the two hierarchies reduces the dimensionality of the learning, avoiding the "big data" training of current ML methods.

Unlike current ML methods, we start from some knowledge embedded in the system. This takes the form of:

- A set of simple model shapes such as line segments, circles, rectangles, boxes, etc., placed at the bottom of the parts hierarchy and at the top of the abstraction hierarchy.

- A set of functions expressing relations between shapes, e.g. distance, angle difference, lengths ratios.

- A set of rules for building the dual hierarchies.

We proceed using processes similar to the recognition algorithm and akin to the scientific method:

- **Generation of grouping hypotheses.** Examining the image examples, we look for instances in which simple objects have recurrent relations (as measured by our relation functions). For example, the system may see two boxes that are at a certain distance and angles relative to each other. (A box is a basic shape that the system has already recognized.) If it sees this arrangement multiple times, it generates a hypothesis of a possible object that it may call a "truck". Statistically, we look for arrangements that are not explainable by a random distribution and hypothesize them to be probable objects.

- **Verification and specification.** The system checks the given images to see if there are other objects similar to the hypothesized "truck". If this happens consistently then the "truck" is considered a verified object. A new node "truck" is added to the model graph. New nodes for "cab" and "trunk" are added as its parts lower in the parts hierarchy, and under the built-in "box" node in the abstraction hierarchy, with the appropriate links (including attributes). It may find circles in certain positions consistently under the boxes that gave rise to the trucks, so new nodes "wheel_i" are added to the model graph under the built-in "circle" node in the abstraction hierarchy, and under the "truck" in the parts hierarchy, with the appropriate links. This is in line with our notion that a generic object such as a circle acquires a specific meaning like a wheel in the context of its group, the truck. We have thus moved bottom-up in the parts hierarchy and top-down in the abstraction hierarchy. The system notes the distributions of the values of the various geometric relation functions involving the parts of an object, and uses them to assign the mean values and elasticity constants of these relations.

In this initial step above, the relations are quite loose and inclusive to fit a high-level abstraction of a truck. In the next iteration, the system learns to differentiate between the specific "truck1" and "truck2". The system looks at several trucks in the images and checks the relations between parts of the trucks, such as distances or size ratios of the cab and trunk. It finds that these quantities do not follow a random distribution, but some relations values occur more frequently than others. It assigns one set of frequent values as "truck1" and the other as "truck2". It adds these nodes under the "truck" node in the abstraction hierarchy, which thus becomes a generic truck. This is in line with our notion that tighter geometric relations between parts lead to more specific objects. The iteration stops when all variations in the image can be explained as random errors.

We have seen that separating the abstraction from the parts hierarchy and using our sources of abstraction described in Section 4 enable the system to build the model graph automatically.

Biologically, one can speculate that certain generic objects are built into the vision system. For instance, a generic "predator" is built into our brains because it would be too late for the victim to learn this from examples. This may be why mythical monsters in movies or cartoons can look even scarier than real creatures even though we have never seen their likes before. Specific predators such as "lion" and "tiger" can be learned later from examples.

## 9. Implementation

The method has been implemented in MATLAB and its GNU clone Octave. The graph traversal algorithm, many geometric relations, basic shapes such as 3D boxes, and other objects such as trucks were tested. The current implementation is still in a prototype stage and we tested it only on a limited set of images.

Several issues arise in implementing the algorithm. Among them:

**Representation of nodes and links.** Most modern programming languages such as C, Fortran and MATLAB, have a data structure called a "structure array". We implement our graph representation as such an array. The structure contains fields that can be accessed by their names, and these can contain quite arbitrary content. In a typical example, the fields contain employee name, address, phone number, etc. The structures are indexed as an array to account for different employees. In our implementation, each node is represented by a structure having fields such as:

| | |
|---|---|
| **Type**: truck | The name ("type") of the object |
| **Instance**: 1 | An object (node) instance number in the image |
| **Parts**: cabin, trunk-a, trunk-b, wheels | Names of part nodes including variants |
| **Groups**: convoy | Names of group nodes the object is linked to |
| **Higher LoA**: vehicle | Names of nodes at a higher level of abstraction |
| **Lower LoA**: truck1, truck2 | Names of nodes at a lower level of abstraction |
| **Origin**: *x,y,z* | Coordinates of object's origin (e.g. center of mass) |
| **Axes**: *ax1,ax2,ax3* | Object's principal axes |
| **Elasticity** *k* | Spring constants for the above quantities |
| **Part(i) origin:** *x,y,z* | Origin of part *i* |
| **Part(i) axes:** *ax1,ax2,ax3* | Principal axes of part *i* |
| **Part(i) elasticity:** *k(i)* | Elasticity constants of part *i* |
| **Midx:** cabin, trunk-a: truck1 | Hypotheses indexed by parts (in model graph) |
| **Relations**: ("size-ratio", cab, trunk-a, 2, 0.3) | Relations between parts |
| ("distance-ratio", cab, trunk-a, 1.5, 0.3) | for hypothesis generation |

The "midx" field lists hypotheses of possible groups, indexed by two parts. These hypotheses are screened using the relations between these parts listed in the "relations" field. These two parts are sufficient to generate a hypothesized

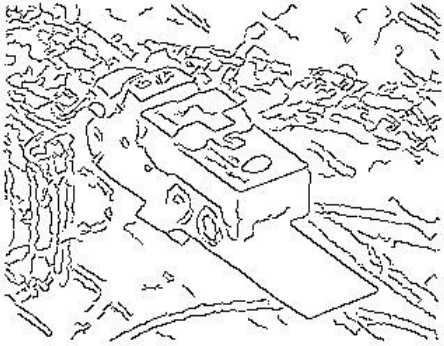
Figure 8: Edge map.

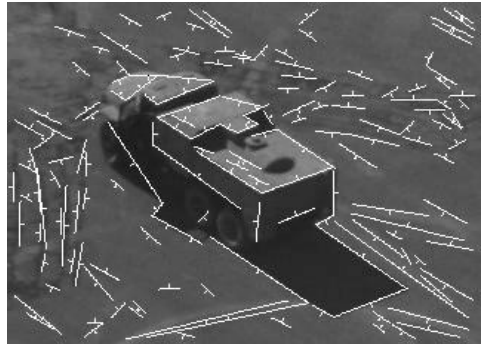
Figure 9: Line segments.

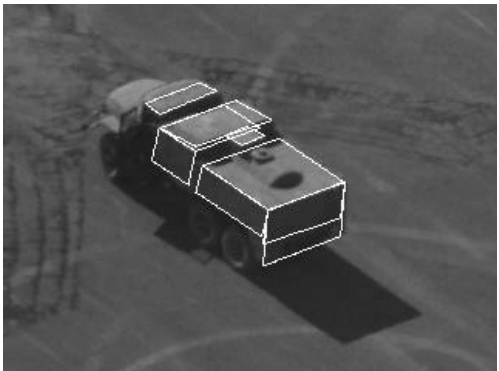
Figure 10: View 1. Rectangles, boxes.

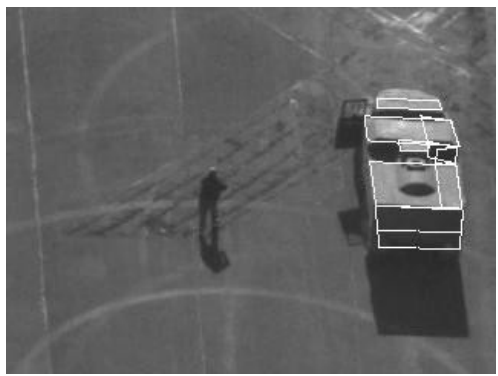
Figure 11: View 2. Same ratios of parallel sides and distances as in View 1.

coordinate transformation between the model and the image. We can then perform verification by transforming the model onto the image and check these and other parts.

The data in the "relations" field is inspired by the syntax of Lisp. Each relation is a list whose first element is a name of a relation function, in this example "size-ratio". Given the hypothesis "truck1", this function finds the ratio of sizes of the "cabin" and the variant "trunk-a" of the trunk. A hypothesis "truck1" proceeds to verification when this relation is satisfied. For this relation to be satisfied, the function "size-ratio" has to return the value of "2" with tolerance 0.3. This function is one of our general relation functions that can be checked. These can be geometric such as "ratio", "pose" (checking relative poses including parallelism), or they can be topological relations such as "touch" or "inside". These generic functions can be used for any objects.

**Low-level features.** Features such as edges are integral parts of the hierarchies. An edge cannot be detected reliably on its own but as a part of a larger object. An edge should be strengthened if it is a part of (e.g.) a line segment and weakened otherwise, and this can be handled naturally by our probability propagation algorithm. However this is not fully implemented yet and for the current examples we have used standard edge detectors.

**Symmetric objects** have multiple representations. A line segment does not have a direction, but it is hard to represent it analytically without a direction, thus it has two representations corresponding to the two possible directions. Similarly, a box has 48 representations. This complicates the testing of the hypotheses as we need to test 48 boxes instead of one. We have found a way to avoid this by using a testing method that is invariant to these different representations.

**Projection from 3D to 2D.** Although a box consists of rectangles in 3D, these rectangles are projected onto 2D as parallelograms (assuming affine projection). The hypothesis generation must be adjusted accordingly. Some relations are invariant to this projection and can be used to screen hypotheses. In [11], such invariant relations were derived for a cloud of points, but they do not take the structure of the object into account. In [10], invariants of curves were used based on derivatives. In the current implementation, we use various algebraic invariants such as the ratios of parallel line segments. For instance, given parallel lines in 2D, we can hypothesize that they are parallel in 3D.

In the following example, we have tested the viewpoint invariance of our algorithm using a 3D truck projected from different points of view. Figure 8 shows an edge map generated by a conventional edge detector, and Figure 9 shows line segments based on these edges. These were fed into our algorithm. Figure 10 shows the resulting boxes and a truck recognized by our algorithm as the "truck1" expressed in the model graph. Figure 11 shows the result of our algorithm applied to the same truck seen from a different viewpoint. We have recognized it as the same truck. This is because it has the same structure and invariant relations in both images. That is, the ratios of lengths of parallel sides of various rectangles, as well as the ratios of sides of rectangles to distances in parallel directions, are the same in both views. There is no 2D geometric transformation between two 2D images projected from 3D, so they cannot be matched by a template-based method even if it tries to account for transformations.

## 10. Conclusion

We have presented an object recognition system based on the structure and geometry of objects, using a machine-learnable model graph. The system is more versatile than previous ones by having two distinct hierarchies of models, both of which can be built by the system: parts and abstraction. Although the concepts of abstraction and semantics seem human, we define them in a computational way that has no reliance on human perception or language. This makes it possible to build a fully automated learning and recognition system. This is done by an algorithm that moves in both hierarchies at the same time, combining both bottom-up and top-down strategies. This reduces the dimensionality of both the learning and recognition processes, avoiding the need for "big data" training.

Because the system classifies specific objects into generic ones based on intrinsic structure rather than pixels, the classification is much more robust than current ML methods with respect to small variations in the object. A stop sign will not be misidentified just because it has some stickers on it. In this way our system is similar to human recognition, and therefore the classification is transparent and understandable by humans, unlike current ML classifications. While the method is specialized for visible objects, its underlying principles have much wider applicability.